%% file: ISRPS_main.tex
\documentclass{isprs} 
\usepackage{subfigure}
\usepackage{setspace}
\usepackage{geometry} 
\usepackage{epstopdf}  
\usepackage[labelsep=period]{caption}  
\usepackage[british]{babel} 
\usepackage[hang]{footmisc}
\usepackage{url}
\usepackage{booktabs}
\usepackage{xcolor}
\usepackage{amsmath}

\pdfoutput=1

\begin{document} 

\title{Scaling Laws for Geospatial Foundation Models: A case study on PhilEO Bench} 

\author{\parbox{\textwidth}{\centering Nikolaos Dionelis\textsuperscript{1}, Riccardo Musto\textsuperscript{2}, Jente Bosmans\textsuperscript{1,3}, Simone Sarti\textsuperscript{2}, Giancarlo Paoletti\textsuperscript{2}, Peter Naylor\textsuperscript{1}, \\ Valerio Marsocci\textsuperscript{1}, Sébastien Lefèvre\textsuperscript{4}, Bertrand Le Saux\textsuperscript{1,5}, Nicolas Longépé\textsuperscript{1}\thanks{Corresponding author}}}

\address{1. European Space Agency (ESA), $\Phi$-lab. \\ 2. Leonardo Labs, Italy. \\ 3. VITO. \\ 4.
IRISA, Université Bretagne Sud. \\ 5. European Commission (EC)}  

\abstract{Foundation Models (FMs) have achieved state-of-the-art performance across domains by leveraging large-scale pretraining. In Earth Observation (EO), the availability of petabyte-scale satellite archives has recently enabled the development of GeoSpatial Foundation Models (GFMs). Yet, fundamental questions remain regarding how dataset size, model architecture, and size interact to determine downstream performance. In this work, we systematically explore this design space by pretraining and fine-tuning models on three dataset scales: PhilEO Globe (0.5TB), FastTOM (2TB, introduced here), and MajorTOM (23TB). We evaluate three architectural families: Geo-Aware U-Net (CNN), ViT-UPerNet (Transformer), and Mamba (State-Space Model); across model sizes ranging from 44M to 300M parameters. All models are benchmarked on PhilEO Bench, covering: road density and building density regression, and land cover segmentation, and are compared against existing GFMs such as TerraMind and Prithvi-EO-2.0.
Our results show that CNN-based models remain highly competitive in low-shot settings, with a 200M-parameter Geo-Aware U-Net outperforming larger architectures on regression tasks. However, when scaling to multi-terabyte datasets, ViT-UPerNet achieves the best performance, particularly for semantic segmentation on MajorTOM (23TB). Finally, we provide the first extensive evaluation of Mamba models in EO, highlighting their potential efficiency advantages, though further large-scale pretraining is required to fully match CNNs and ViTs.
All code, pretrained models, and the FastTOM dataset are released publicly, enabling reproducibility and further exploration of scaling laws for GFMs.}



\maketitle

\input{1_introduction}

\input{2_background}

\input{3_results}

\input{4_discussionconclusion}

{
	\begin{spacing}{1.17}
		\normalsize
		\bibliography{ISPRSguidelines_authors} 
	\end{spacing}
}
\input{5-annexMamba}

\end{document}

%% file: 1_introduction.tex
\section{Introduction}\label{sec:intromainsec}       

Foundation Models (FMs) have recently emerged as a central paradigm in Artificial Intelligence (AI) \cite{bommasani2021opportunities}, shifting the focus from training specialized models for individual tasks to building general-purpose models that can be adapted across tasks. These models are trained on vast datasets, contain billions of parameters, and leverage self-supervised learning (SSL) strategies. The most prominent success stories come from Large Language Models (LLMs) \cite{brown2020language}, which have shown how large-scale training on diverse corpora can yield highly transferable representations.

This paradigm has rapidly expanded to other domains where massive datasets are available. In Earth Observation (EO), the sheer volume of satellite data makes the field particularly well-suited for Foundation Models. For instance, the Copernicus Sentinel-2 constellation alone produces approximately 1.6TB of imagery per day, enabling the training of large-scale Geospatial Foundation Models (GFMs).

GFMs are deep neural networks trained with SSL not for a single downstream task, but for learning broadly useful geospatial representations. The effectiveness of such models depends critically on two main factors: (i) the choice of architecture and training protocol, which determines how well the model can encode spatial and temporal structure, and (ii) the scope and quality of the training data, given that not all EO data is usable, and dataset design choices strongly affect model generalization.

Compared to generic vision foundation models, GFMs must capture domain-specific properties of EO data. Key challenges include: (a) \textbf{spatio-temporal awareness}, as EO signals evolve across space and time; (b) \textbf{multi-scale reasoning}, since relevant phenomena range from fine-grained objects (e.g. buildings) to global processes (e.g. climate zones); (c) \textbf{sensor agnosticism}, because EO datasets are heterogeneous in resolution and modality; and (d) \textbf{cross-modality integration}, to exploit complementary information from optical, radar, climate projections, or other sources.

To achieve meaningful geospatial encodings, one must therefore carefully explore the design space of GFMs: which architectures are most suitable (CNNs, Transformers, or State Space Models), how model size interacts with dataset scale, what pretraining strategies are effective, and how to benchmark progress. This also raises a fundamental question: do GFMs follow scaling laws similar to those reported for LLMs \cite{kaplan2020scaling}, where larger datasets and models consistently improve performance, or are there domain-specific bottlenecks that break this trend?

\textbf{Our contribution.} In this study, we address these questions by systematically exploring the design space of GFMs using the PhilEO Bench \cite{phileo}. Our contributions are fourfold:
\begin{itemize}
    \item[(i)] We release a downscaled version of the MajorTOM dataset, named {FastTOM}, enabling faster experimentation prior to scaling to the full 23TB corpus;
    \item[(ii)] We pretrain and fine-tune models across three dataset scales: small (0.5TB), medium (2TB), and large (23TB); and systematically vary model depth (from 44M to 300M parameters) and architecture (Geo-Aware U-Net, ViT-UPerNet, and Mamba);
    \item[(iii)] We evaluate these models on PhilEO Bench \cite{phileo}, made of three tasks with Sentinel-2 data (i.e. road and building density regression, and land cover segmentation), so to standardize the evaluation protocol and draw some consistent conclusions on outcomes; 
    \item[(iv)]  We measure computational efficiency (FLOPs) \cite{paperforflops} and analyze trade-offs between dataset scale, model complexity, and downstream performance.
\end{itemize}

All code and pretrained models are released publicly:  
\url{http://github.com/ESA-PhiLab/PhilEO-MajorTOM}.  
Our results indicate that large GFMs trained on global datasets can transfer effectively even to tasks requiring only a subset of the encoded information, thus supporting their broader applicability in EO pipelines.

\section{Related Work}\label{sec:backgroundsec}      

\subsection{Datasets}  
One of the key ingredients to GFM is data. 
Because of the number of sensors, but also the number of downstream tasks, large datasets have been created and curated with the purpose of training large Deep Neural Networks (DNN).
EO datasets have been found to focus on multi-spectral satellite imaging, to incorporate multi-sensors and multi-source data, and to be tailored to specific tasks like object detection. 
Cherry picking, and selecting the training data scope (size, source, sensors, ...) fit within the design of GFM and is a crucial step.
As of today, many large datasets have been released.
The Global Harmonized Landsat Sentinel-2 (HLS) dataset fused Landsat and ESA Sentinel-2 imagery by reducing the original data to a common denominator \cite{newibmnasa}. 
The PhilEO Globe \cite{phileo}, provided with the PhilEO Bench, is a small-ish (0.5TB) dataset of ESA Sentinel-2 imagery with annotations on roads, building, and land cover classification.
Specialised for SSL training, the SSL4EO-S12 \cite{wang2023ssl4eo} consists of ESA Sentinel-1/2 data with a seasonality component.
Copernicus-Pretrain was proposed, and consists of ESA Sentinel, from 1 to 5P \cite{TUMYiWangCopFM}.
Functional Map of the World (fMoW) \cite{fmnewsatmaepp} is a large annotated dataset (with bounding boxes like annotations) from the Maxar constellation and consists of 4 to 8 bands.
TerraMesh \cite{blumenstiel2025terramesh} is  a large multi-modal dataset that includes, Sentinel-1, 2 and land cover classification maps, topography maps and NDVI maps.
Finally, MajorTom \cite{mtom} is the largest existing ESA Sentinel-2 dataset that provides one cloud-free snapshot of every location on earth, including imagery of the oceans and land ice.
In Table \ref{tab:maintableliter}, we recap these large datasets.

\begin{table}[!tbp]
\caption{\centering Comparison of multispectral pretraining datasets for GFMs. Distinctive features are highlighted.}
\centering
\label{tab:maintableliter}
\vskip -0.0706in
\begin{flushleft}
\begin{scriptsize}
\begin{sc}
\begin{tabular}{p{2.8cm} p{0.6cm} p{1.cm} p{2.2cm}}
\toprule
\textbf{Dataset} & \textbf{Size} & \textbf{Coverage} & \textbf{Feature} \\
\midrule
Global HLS \cite{newibmnasa} & 4–5TB & Land & HLS time series \\
\midrule
fMoW \cite{fmnewsatmaepp} & 3.5TB & Land & High-res objects \\
\midrule
SSL4EO-S12 \cite{wang2023ssl4eo} & 1.5TB & Land & Multi-temporal S2 \\
\midrule
Copernicus-Pretrain \cite{TUMYiWangCopFM} & 10TB & Land & Large S2 archive \\
\midrule
TerraMesh \cite{PaperTerraMind} & 14TB & Land, ocean, ice & Multi-modal global \\
\midrule
PhilEO Globe \cite{phileo} & 0.5TB & Land & Seasonal S2, 4x/year \\
\midrule
\midrule
{MajorTOM} \cite{mtom} & {23TB} & {Land, ocean, ice} & {Full-scale S2 global} \\
\midrule
\bottomrule
\end{tabular}
\end{sc}
\\
\vskip 0.05in
HLS = Harmonized Landsat–Sentinel-2, S2 = Sentinel-2, fMoW = Functional Map of the World.
\end{scriptsize}
\end{flushleft}
\vskip -0.1in
\end{table}

\subsection{Geospatial Foundation Model benchmarks}

The rise of Geospatial Foundation Models (GFMs) has brought to the creation of benchmarks to evaluate their performance across diverse tasks. While valuable progress has been made, most of the benchmarks often focus on narrow domains (e.g., agriculture or forestry), on limited tasks (e.g., classification without change detection), or overlooking key EO characteristics such as multi-temporality and multi-modality.

Some efforts include FoMo-Bench \cite{fomobench}, targeting forest monitoring, and Chang et al. \cite{Chang2024OnTG}, focused on agricultural applications across multiple continents.

At a broader scale, GEO-Bench \cite{geobench} and SustainBench \cite{sustainbench} attempt to unify diverse datasets across modalities, resolutions, and sustainable development goals. Similarly, EarthNets \cite{earthnets} compiles a variety of published datasets, but does not provide an easy entry point to use them.

PhilEO Bench \cite{phileo} is particularly well-suited for Sentinel-2 data, providing dense annotations for regression tasks, like building and road density and for segmentation tasks with land cover classification. This makes it an excellent benchmark for evaluating models like MajorTOM that are explicitly pretrained on Sentinel imagery. 

PANGAEA \cite{pangaea} represents another step toward large-scale, standardized evaluation of GFMs. It spans a diverse set of EO tasks, emphasizing not only performance but also robustness and transferability across domains. Yet, its scope is broader than targetting scalability, meaning specialized datasets like PhilEO remain indispensable for Sentinel-focused evaluation, under controlled conditions.

\subsection{Geospatial Foundation Models}

GFMs are pretrained on large-scale satellite data to learn transferable feature representations, which can then be fine-tuned on smaller labeled datasets for a variety of downstream tasks such as classification, regression, and segmentation \cite{phileo}. Architectures for GFMs have evolved from CNN-based models such as U-Net, still a strong baseline for semantic segmentation, to transformer-based approaches (ViT, Swin) that better capture long-range dependencies, albeit with higher computational cost \cite{Longepe_2025}. More recently, state-space models (e.g., Mamba) have been proposed to achieve scalable global modeling with linear complexity.

Training GFMs at scale relies heavily on self-supervised learning (SSL). Methods such as Masked Auto-Encoders (MAE) \cite{MAE}, contrastive learning \cite{CL}, diffusion models \cite{diffusion}, and Joint Embedding Predictive Architectures (JEPA) \cite{paperforjepa1} have proven effective for exploiting massive unlabeled datasets. These approaches share the core hypothesis that SSL reduces the need for costly labeled data by leveraging global EO corpora. Recent advances, such as Meta’s DINOv3 \cite{simeoni2025dinov3}, further demonstrate that large-scale SSL pretraining on hundreds of millions of satellite images enables state-of-the-art performance across diverse geospatial tasks.

Despite these advances, most existing GFMs are still limited by modest dataset scales (often below 20 TB), with performance varying significantly across applications. In some cases, conventional models remain competitive. This raises an open question for the field: should the focus be on a single universal GFM, on task-specific architectures, or on strategies such as Mixture of Experts (MoE)? To address this gap, we pretrained and assess a diverse set of architectures on different sizes of datasets.

Recent studies, such as \cite{cromapaper,pangaea,PaperTerraMind,deflect}, confirm the potential of GFMs, while also showing that U-Nets remain strong baselines and that transformer variants like Swin-UPerNet can rival them in certain tasks \cite{upernet5}. Furthermore, few-shot learning has emerged as a promising direction for GFMs, enabling models to generalize effectively from limited labels \cite{nshot1,nshot2,nshot3,nshot4}.




%% file: 2_background.tex

\section{Architectural Families for Geospatial Foundation Models}\label{sec:propmodel} 

The development of GFMs is fundamentally driven by pretraining. By leveraging massive corpora of EO data, models acquire transferable representations that can later be fine-tuned for specific applications, often requiring only a small fraction of labeled data. The choice of architecture directly impacts both the type of features a model can learn and the scale of data required for effective pretraining. In this section, we examine three architectural families: CNNs, Vision Transformers, and State-Space Models (Mamba); highlighting their methodological differences, strengths, and limitations in the context of GFMs.

\subsection{CNN Models: Geo-Aware U-Net}

Convolutional Neural Networks (CNNs) are characterized by strong inductive biases that make them well-suited for imagery. The U-Net, in particular, has been the de facto standard for semantic segmentation in EO due to its encoder–decoder design with skip connections, which preserves spatial detail while capturing abstract representations.

The Geo-Aware U-Net introduced and evaluated on the PhiLEO Bench \cite{phileo} extends this classical architecture by incorporating three pretext tasks during pretraining: (i) masked reconstruction of input patches, (ii) geo-location estimation, and (iii) Köppen-Geiger climate zone prediction. This “geo-awareness” ensures the model does not simply memorize textures but learns geographic and climatic context as well. When pretrained with the different amount of Sentinel-2 data, this model demonstrated strong generalization to downstream tasks, even when a small quantity of data was used.

The main advantage of CNNs lies in their efficiency and spatial precision. They require less data to achieve good performance compared to Transformers, making them highly valuable in settings where pretraining is limited to sub-terabyte datasets. However, their locality-driven convolutions inherently restrict their ability to model global correlations across large spatial extents. This limitation becomes more evident when scaling to multi-terabyte corpora, where their representational capacity may saturate.

\subsection{ViT Models: PhilEO ViT-UPerNet}

Vision Transformers (ViTs) mark a shift from local convolutions to global self-attention. In theory, every token can attend to every other token, enabling the model to capture long-range dependencies and global patterns that are essential for many EO tasks, such as large-scale land cover mapping or ocean monitoring.

The ViT-UPerNet introduced and evaluated on the PhilEO Bench, \cite{phileo} couples a ViT backbone with the UPerNet decoder, which fuses hierarchical features across scales and enhances global context with a pyramid pooling module. For this study, we pretrain this architecture with different datasets scales: small, medium and large.  The scale of data is critical: while ViTs often underperform CNNs when trained on smaller corpora, their ability to model global context begins to dominate once exposed to tens of terabytes of diverse imagery \cite{vit}.

The strength of ViTs lies in their flexibility and global receptive field. They can outperform CNNs on large-scale benchmarks, provided sufficient pretraining data and computational resources. However, this comes with significant costs: self-attention scales quadratically with sequence length, making them computationally demanding, and they exhibit weaker inductive biases than CNNs, meaning they depend heavily on large datasets to avoid overfitting or underutilizing their capacity.

\subsection{State-Space Models: Mamba} \label{ssec:mamba}

State-Space Models (SSMs), and in particular the recent Mamba S6 \cite{papermamb6}, offer a third pathway. Instead of self-attention, Mamba employs a selective scanning mechanism that enables modeling of long-range dependencies with linear complexity relative to sequence length. This makes it highly attractive for EO imagery, which often involves high-resolution, wide-coverage data.

In the context of this work, we adapted Mamba into 2D architectures that preserve spatial continuity \cite{papermamb7}, avoiding the distortions introduced by naive 1D sequence unrolling. These adaptations ensure that local neighborhoods in satellite images are respected, while still enabling efficient capture of long-range spatial patterns. While Mamba is still relatively new in EO applications, its potential lies in offering a scalable middle ground between CNNs and Transformers: it retains stronger inductive bias than ViTs, yet scales more efficiently to global contexts.

The main drawback is its lack of maturity. Unlike CNNs and Transformers, which have extensive benchmarks and tooling, Mamba models are still experimental in geospatial domains. Their performance strongly depends on carefully designed pretraining pipelines, and the lack of large-scale EO benchmarks specifically tailored to SSMs makes systematic evaluation more challenging.

\subsection{Comparative Summary}

In summary, CNNs, ViTs, and Mamba represent three distinct philosophies for geospatial modeling. CNNs excel at efficiency and local detail, making them ideal for smaller datasets (hundreds of GBs to $\sim$1TB). ViTs, while resource-hungry, can fully exploit massive datasets (10–20TB and beyond), achieving superior global modeling. Mamba, though newer, promises linear scalability and may bridge the gap between efficiency and global context, provided adequate pretraining. Across all three families, the decisive factor is the scale and diversity of pretraining data: smaller datasets may favor CNNs, while multi-terabyte corpora unlock the potential of ViTs and Mamba.

\begin{table}[ht]
\centering
\caption{Comparison of GFM Architectural Families.}
\label{tab:gfm_arch_comparison_short}
\begin{tabular}{p{1.8cm}p{2.7cm}p{2.8cm}}
\toprule
\textbf{Arch.} & \textbf{Pros} & \textbf{Cons / Data Needs} \\
\midrule
\textbf{CNN} (U-Net) & Strong spatial bias; Efficient; Good with limited data. & Limited global context; Saturation at $\sim$1–2TB. \\
\midrule
\textbf{ViT} (UPerNet) & Global context; Flexible tasks; SOTA at scale. & Quadratic cost; Weak bias; Requires $\geq$10–20TB. \\
\midrule
\textbf{SSM} (Mamba) & Linear scalability; High-res imagery; Better EO bias than ViT. & Experimental; Limited benchmarks; Needs multi-TB pretraining. \\
\bottomrule
\end{tabular}
\end{table}

%% file: 3_results.tex
\section{Evaluation and Results}\label{sec:evalsec}                                 

\subsection{Evaluation Protocol}

Our experiments evaluate model performance across varying pretraining data scales, architectures, and downstream tasks. We consider three pretraining corpora of satellite Sentinel-2 multi-spectral data of increasing size (0.5TB, 2TB, and 23TB), which represent small-, medium-, and large-scale Earth observation scenarios. Specifically:  
\begin{itemize}
    \item GlobeEO 0.5TB \cite{phileo}, made of S2L2A images that focus only on land, \textit{excluding} oceans and ice. For each location, it includes four time steps, one per season.
    \item MajorTOM 23TB \cite{mtom}: a subset of the MajorTOM Core-S2L2A, made of approximately 60TB of unlabeled global data, covering most of the Earth, including oceans and ice. 
    \item FastTOM 2TB: a subset of the previous MajorTOM 23TB
    , containing only land and excluding oceans and ice, making it more task-specific for terrestrial downstream tasks.  
\end{itemize}

To study generalization and efficiency we compare three families of architectures:
\begin{itemize}  
  \item U-Net, where we examine both trained from scratch and pretrained, the latter followed the pretraining protocol presented in \cite{phileo} and is named Geo-Aware U-Net.
The model is pretrained using a multi-objective loss, combining image reconstruction (similar to masked autoencoding) and geo-location prediction (longitude and latitude estimation), each with dedicated output heads. Leveraging the U-Net CNN architecture, this model is particularly effective at capturing local spatial correlations in EO imagery. We included different flavours of models, based on the different number of parameters (i.e. 44M and 200M) and amount of pretraining data, that follow this same protocol.
  \item ViT: a transformer-based backbone of different scales (i.e. 100M, 200M and 300M) and used with different decoders (i.e. ViT-UperNet, ViT-CNN).
  \item Mamba: a state-space model tailored for long-sequence processing. We included a general version (i.e. Mamba UPerNet) and a version tailored on RS data (i.e. RS3Mamba).
\end{itemize}
All the available models are introduced in Table \ref{tab:maintableevalseeeccc}.
Experiments are presented in a staged progression (small, medium, large) that first informs model selection, then contrasts architectures, and finally evaluates large-scale pretraining and benchmark performance.

All models are pretrained on Sentinel-2 multispectral imagery, following the protocol presented in \cite{phileo}. Inputs are 128×128 tiles with 10 spectral bands (i.e. the Sentinel-2 20m bands and 60m band are resampled to 10m)\footnote{\url{http://sentiwiki.copernicus.eu/web/s2-mission##S2Mission-SpatialResolutionS2-Mission-Spatial-Resolutiontrue}}. While this resampling creates a spatially consistent 10m input tensor, it introduces a trade-off: it prevents the model from leveraging the unique spectral signatures at their native resolutions, and the downsampled pixels represent a spatial average rather than a true 10-meter measurement.

The aforementioned models and pretraining strategies are then evaluated on three relevant downstream tasks using Sentinel-2 data, that consist in the PhilEO benchmark \cite{phileo}. The Sentinel-2 satellite constellation provides globally consistent multi-spectral data at a scale, resolution and are freely available that makes it the \textit{de facto} standard for EO foundation models. Specifically:
\begin{itemize}   
    \item \textbf{Building density estimation} is a pixel-wise regression downstream task, in which building density is the percentage of the area that is covered by buildings. The task takes into account the building cover and does not account for the building height. 
    \item \textbf{Road density estimation} is a pixel-wise regression downstream task. Similar to the previous except that this downstream task estimates how much of the area is covered by roads.        
    \item \textbf{Land cover mapping} is a semantic segmentation task with 11 classes, using the ESA WorldCover dataset. The categories are: Tree cover, Shrubland, Grassland, Cropland, Built-up, Bare/sparse vegetation, Snow and ice, Permanent water bodies, Herbaceous wetland, Mangrove, and Moss and Lichen\footnote{http://worldcover2020.esa.int/data/docs/WorldCover$\_$PUM$\_$V1.1.pdf}.           
\end{itemize}
Evaluating on PhilEO Bench spans both dense regression and segmentation tasks across urban and environmental domains. Together, these tasks form a compact but representative benchmark to test generalization and efficiency in GFMs.  

We evaluate the models using an $n$-shot evaluation protocol where we assess the sample efficiency of GFMs in the downstream tasks.
In the figures,  
each point on the $x$-axis represents the number of samples per country. For example, at $n = 50$, we use $6$ countries × $50$ samples = $300$ data samples.
For evaluating pixel-wise regression downstream tasks, we use the Root Mean Squared Error (RMSE). For semantic segmentation, we use Overall Accuracy to not be biased by class distribution.   

Table~\ref{tab:maintableevalseeeccc} summarizes the evaluated models, key characteristics, estimated computational cost (measured in FLOPs), and downstream performance. Final model selection accounts for both task performance and computational cost.

\begin{table*}[!tbp]                                                               
\caption{ \centering Summary of Trained and Evaluated EO Foundation Models and Their Key Characteristics. The best performance per column is in bold, while the second best is underlined (considering $\pm 3 \% \text{ variability})$. GC stands for Group Channel and P16 refers to a patch size of 16.}      
\vspace{8.6pt}\centering              
\label{tab:maintableevalseeeccc}   
\vskip -0.0706in      
\vskip -0.0706in 
\vskip 0in
\begin{center}
\begin{scriptsize}   
\begin{sc}         
\begin{tabular}          
{p{1.3cm} p{3cm} p{2.5cm} p{1.8cm} p{1.3cm} p{2.7cm} p{2.7cm}}      
\toprule                
\midrule    
\textbf{Models} & \textbf{Main features} & \textbf{Particularities} &  \textbf{Versions} & \textbf{FLOPs (G)} & \multicolumn{2}{c}{\textbf{Average Performance}} \\ 
& & & &   &     \multicolumn{2}{c}{Building / Road / Land cover}    \\
& & & &   &     \multicolumn{2}{c}{RMSE $\downarrow$ / RMSE $\downarrow$ / Acc. $\uparrow$}    \\
& & & &    &    limited n-shot \par ($n\leq 100$)  & large n-shot \par ($n\geq 1000$) \\
\midrule
\midrule         
Geo-Aware U-Net & Modified U-Net architecture, Pretrained using reconstruction and geo-location longitude and latitude estimation & Local correlations   & 44M-0.5T   \par 44M-2T  \par 44M-23T \par 200M-2T & $\textbf{5.87}$ \par $82.23$ \par $\textbf{5.87}$ \par $\textbf{5.87}$  & $\underline{0.087}$ / $0.074$ / $\textbf{0.605}$ \par $\textbf{0.076}$ / $\textbf{0.067}$ / $0.548$ \par $\underline{0.087}$ / $\underline{0.072}$ / $0.553$ \par $\underline{0.084}$ / $\underline{0.071}$ / $0.515$ & $\underline{0.059}$ / $0.064$ / $\textbf{0.730}$ \par $\underline{0.061}$ / $\textbf{0.060}$ / $\underline{0.691}$ \par $0.063$ / $\textbf{0.061}$/ $\underline{0.696}$ \par $0.066$ / $\underline{0.062}$ / $0.662$  \\  
\midrule        

ViT with UPerNet decoder & Transformer backbone, Global correlations, Pretrained on 0.5TB, 2TB and 23TB datasets & Multi-scale features using UPerNet decoder  &  300M-0.5T  \par 100M-2T \par 100M-23T \par P16 100M-2T & $532.88$ \par $388.51$ \par $388.51$ \par $92.43$    & $\underline{0.085}$ / $\underline{0.070}$ / $0.575$ \par $\underline{0.086}$ / $\underline{0.072}$ / $0.554$ \par $0.093$ / $\underline{0.072}$ / $0.535$ \par $0.145$ / $0.099$ / $0.430$    & $\underline{0.060}$ / $\textbf{0.060}$ / $\textbf{0.725}$ \par $0.070$ / $0.067$ / $0.647$ \par $0.066$ / $0.064$ / $0.684$ \par $\textbf{0.043}$ / $0.067$ / $0.664$    \\             
\midrule                              
               
ViT with CNN decoder &  Transformer backbone \cite{phileo}, Global correlations, Large architecture & Pretrained on the GlobeEO 0.5TB dataset  & 300M-0.5T \par GC 200M 0.5T & $507.53$ \par $488.23$ & $0.150$ / $0.146$ / $0.445$ \par $0.215$ / $0.200$ / $0.495$ & $0.087$ / $0.080$ / $\underline{0.695}$ \par $0.088$ / $0.078$ / $\underline{0.695}$       \\                    
\midrule           

RS3Mamba \cite{paperRS3Mamba} & Mamba SSM architecture, Global correlations & More efficient than ViT, Linear complexity rather than quadratic &   168M-2T & $18.64$  & $0.127$ / $0.122$ / $\underline{0.584}$    & $0.071$ / $0.066$ / $\underline{0.710}$    \\   
\midrule      

Mamba UPerNet  \cite{papermamb7} & Mamba architecture, Patch size $16$     & Larger patch size, $16$ compared to $4$     & P16 100M & $\underline{12.68}$ & $0.186$ / $0.136$ / $0.456$  & $0.075$ / $0.075$ / $0.648$       \\    
\midrule                 
TerraMind \cite{PaperTerraMind}   & State-of-the-art existing GFM, Correlation learning, Multi-modal    & Very good/ top performance, Pretraining modalities include S-2, S-1, DEM, geo-location, LULC labels, and NDVI   &   v1.0-B \par v1.0-L &  $16.62$ \par $45.72$ & $0.093$ / $\underline{0.072}$ / $\underline{0.598}$ \par $0.093$ / $\underline{0.072}$ / $\textbf{0.621}$    & $0.092$ / $0.071$ / $\textbf{0.727}$ \par $0.092$ / $0.071$ / $\textbf{0.743}$       \\            

\midrule                   
Prithvi-EO \cite{newibmnasa}      & Recent GFM, Pretrained on HLS data ($30$m res., $6$ bands)     & Large architecture: ViT-H (Huge), 600M, Pretrained on land-only dataset  & v2.0 600M TL & $317.99$ & $0.092$ / $\underline{0.071}$ / $0.539$      & $0.092$ / $0.071$ / $\underline{0.706}$      \\              
\midrule                           
\bottomrule            
\end{tabular}    
\end{sc}             
\end{scriptsize}                
\end{center}                     
\vskip -0.1in       
\end{table*}

\input{31_smallscale}
\input{32_architecture}

\input{33_scaling}

\input{34_FLOPS}

\input{35_comparisons}


%% file: 31_smallscale.tex
\subsection{Preliminary Evaluation on Limited-Scale Dataset (0.5 TB)}\label{sec:smallscale}

We begin by pretraining and evaluating architectures on the GlobeEO 0.5TB dataset \cite{phileo}. These experiments are intended to identify general patterns and the most promising architectures before scaling to larger pretraining datasets. The models compared include a pretrained Geo-Aware U-Net (44M parameters), a U-Net trained from scratch, a ViT with an UPerNet decoder, and two ViTs with CNN decoders, i) a ViT-CNN with 300M parameters and ii) a ViT-CNN with Group Channel (ViT CNN GC). In this small-scale setup, we employ the lighter 44M Geo-Aware U-Net to enable fast benchmarking; later, we scale this model up to 200M parameters.

\begin{figure}[!t]     
\centering  
\includegraphics[width=3.45in]{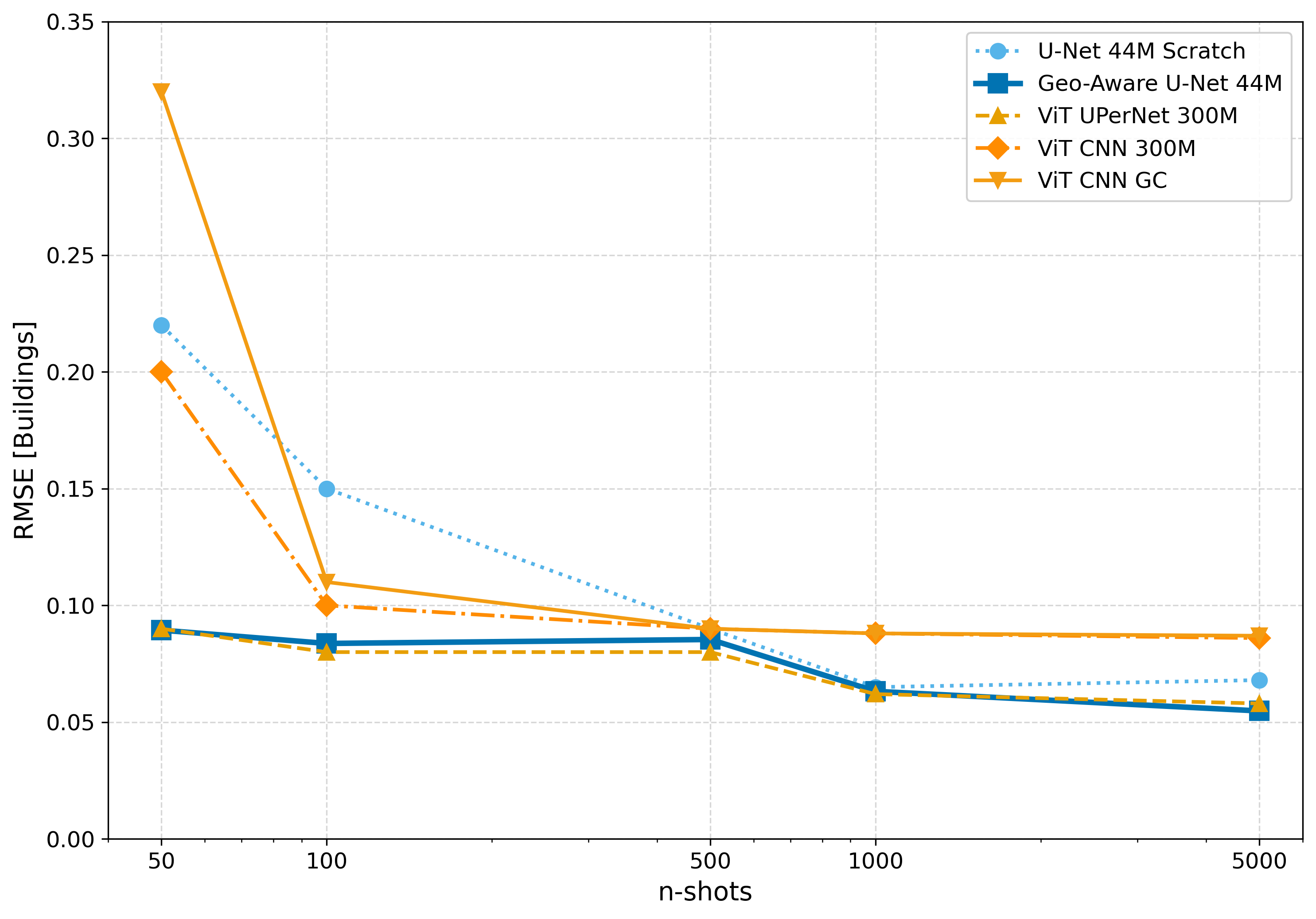}     
\includegraphics[width=3.45in]{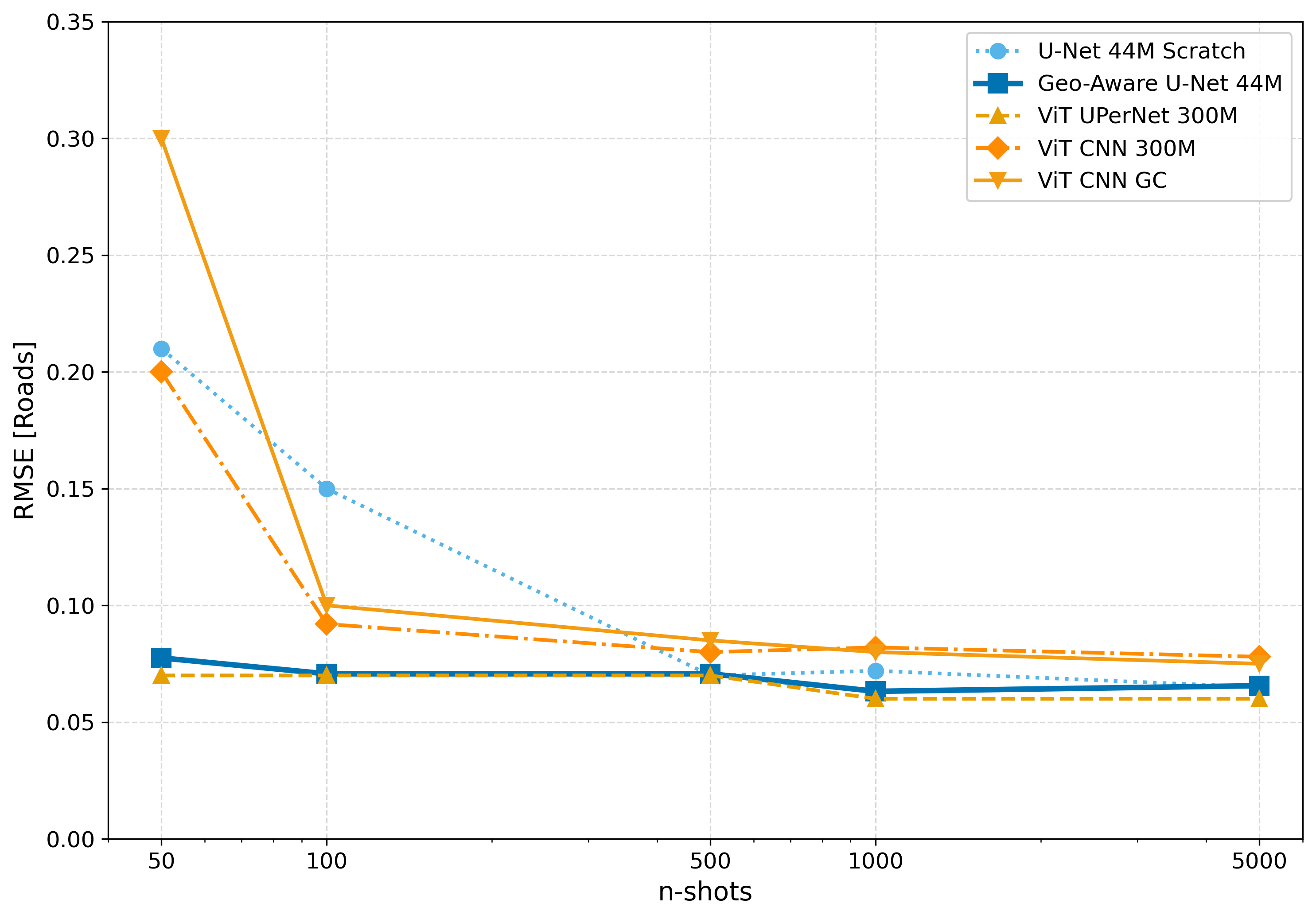}    
\includegraphics[width=3.45in]{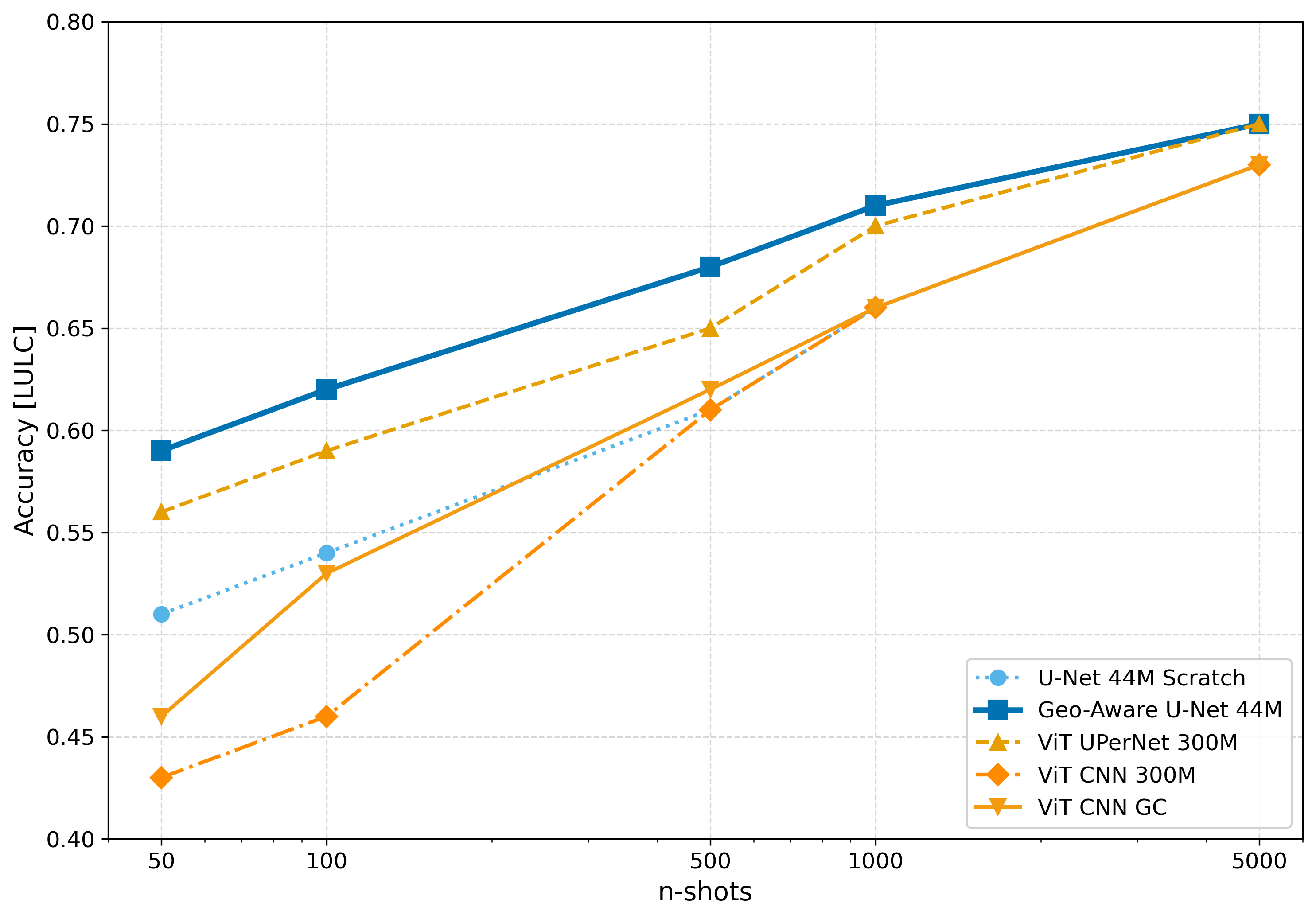}
\caption{Performance on PhilEO benchmark tasks for models pretrained on GlobeEO 0.5TB: Building density regression (top), Road density estimation (middle), and Land cover mapping (bottom), across different $n$-shot settings, where the key takeaways are that pretraining helps, i.e. Geo-Aware U-Net 44M versus U-Net 44M (Scratch), and the UPerNet decoder helps, i.e. ViT UPerNet compared to ViT CNN, where the latter means that a CNN decoder is used rather than multi-scale features in the UPerNet decoder.} 
\label{fig_05TB} 
\end{figure}

Figure~\ref{fig_05TB} and Table~\ref{tab:maintableevalseeeccc} summarize the results on PhilEO Bench. For building and road density estimation, the ViT UPerNet 300M consistently outperforms the ViTs with CNN decoders, and achieves performance comparable to the Geo-Aware U-Net 44M. For example, in the road density task with $n=1000$ shots, ViT UPerNet 300M achieves an RMSE of $0.060$ compared to $0.063$ for U-Net 44M, while both substantially outperform the ViTs with CNN decoders ($\approx 0.080$). This underscores the strength of the UPerNet decoder for pixel-wise regression, which leverages multi-level latent representations.  

Both ViT UPerNet 300M and Geo-Aware U-Net 44M also perform well under limited supervision. On the road downstream task at 50 shots, U-Net 44M achieves an RMSE of $0.078$ while ViT UPerNet 300M reaches $0.070$, showing robustness even in low-shot regimes. The U-Net benefits from skip connections and encoder–decoder structure with convolutional layers, while ViT UPerNet complements this by capturing global context through Transformers together with the multi-scale features of its decoder.

In land cover mapping, both architectures also stand out. At 1000 shots, U-Net 44M achieves an accuracy of $0.71 \%$, while ViT UPerNet 300M yields $0.70 \%$. These results confirm their strong generalization ability for dense semantic segmentation with multiple classes.

Looking across tasks, we also observe that land cover mapping scales more smoothly with the number of shots and reaches higher absolute accuracy, while road and building density estimation are more challenging. In particular, their RMSE values remain higher overall, and the improvements from adding more samples are less pronounced. This suggests that while both U-Net and ViT UPerNet transfer well to semantic segmentation (LULC), fine-grained density estimation tasks such as roads and buildings estimation remain more sensitive to supervision and model design \cite{xxzbuildings}.

Overall, on the 0.5TB pretraining experiments, ViT UPerNet and the Geo-Aware U-Net emerge as the most competitive baselines, consistently outperforming CNN-decoder ViTs and U-Nets trained from scratch across all $n$-shot settings. Because both model size and pretraining dataset scale play important roles, in the next section we increase the capacity of the Geo-Aware U-Net to 200M parameters (see Sec.~\ref{sec:archicomp}).

%% file: 32_architecture.tex
\subsection{Comparative Analysis of Models Pretrained on 2TB Dataset}\label{sec:archicomp}

After evaluating models pretrained on the GlobeEO 0.5TB dataset, we scale-up the pretraining phase using the FastTOM 2TB dataset. This intermediate step allows us to explore how increased data volume impacts model performance before moving to the full-scale MajorTOM 23TB dataset.

To better understand scaling patterns, we extend the analysis to the most promising architectures identified in earlier experiments. This step is crucial for assessing how each model architecture benefits from additional pretraining data and EO content, and whether the performance gains justify the increased computational cost. The models included in this comparison are the Geo-Aware U-Net (44M and 200M parameters), ViT UPerNet 100M, Mamba UPerNet discussed in Section \ref{ssec:mamba}, and RS3Mamba \cite{paperRS3Mamba}.

\begin{figure}[!t]                         
\centering             
\includegraphics[width=3.45in]{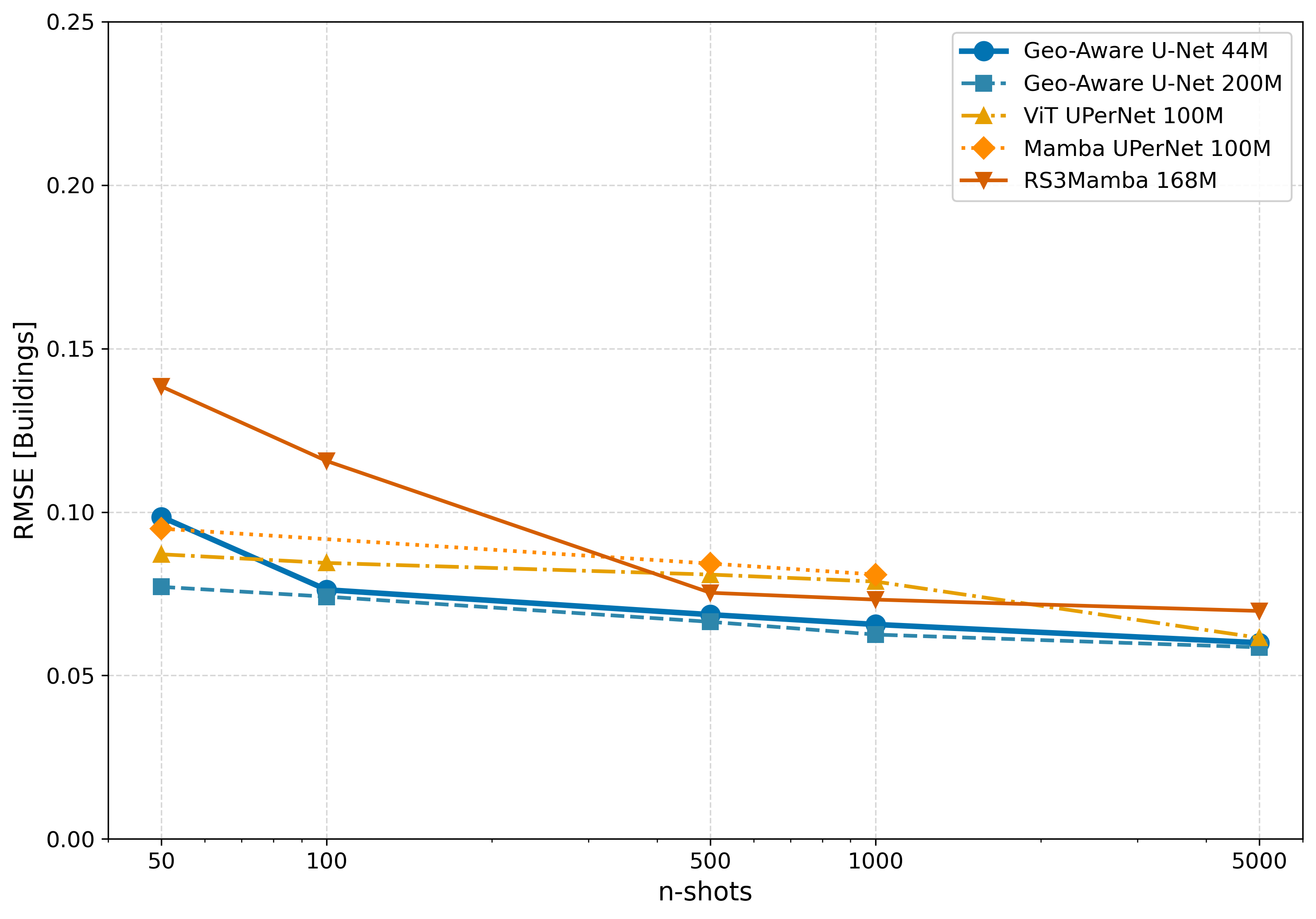}     
\includegraphics[width=3.45in]{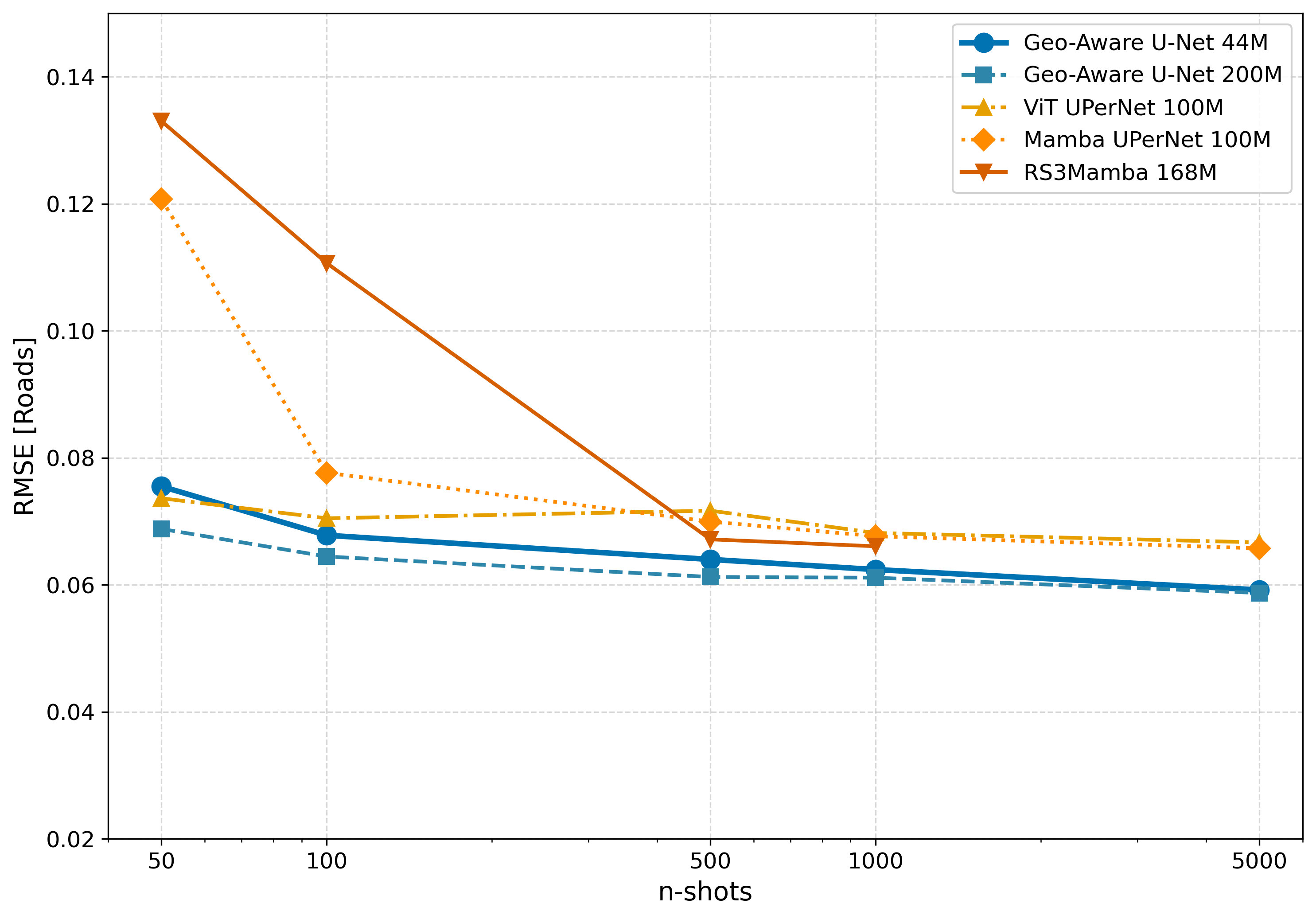}    
\includegraphics[width=3.45in]{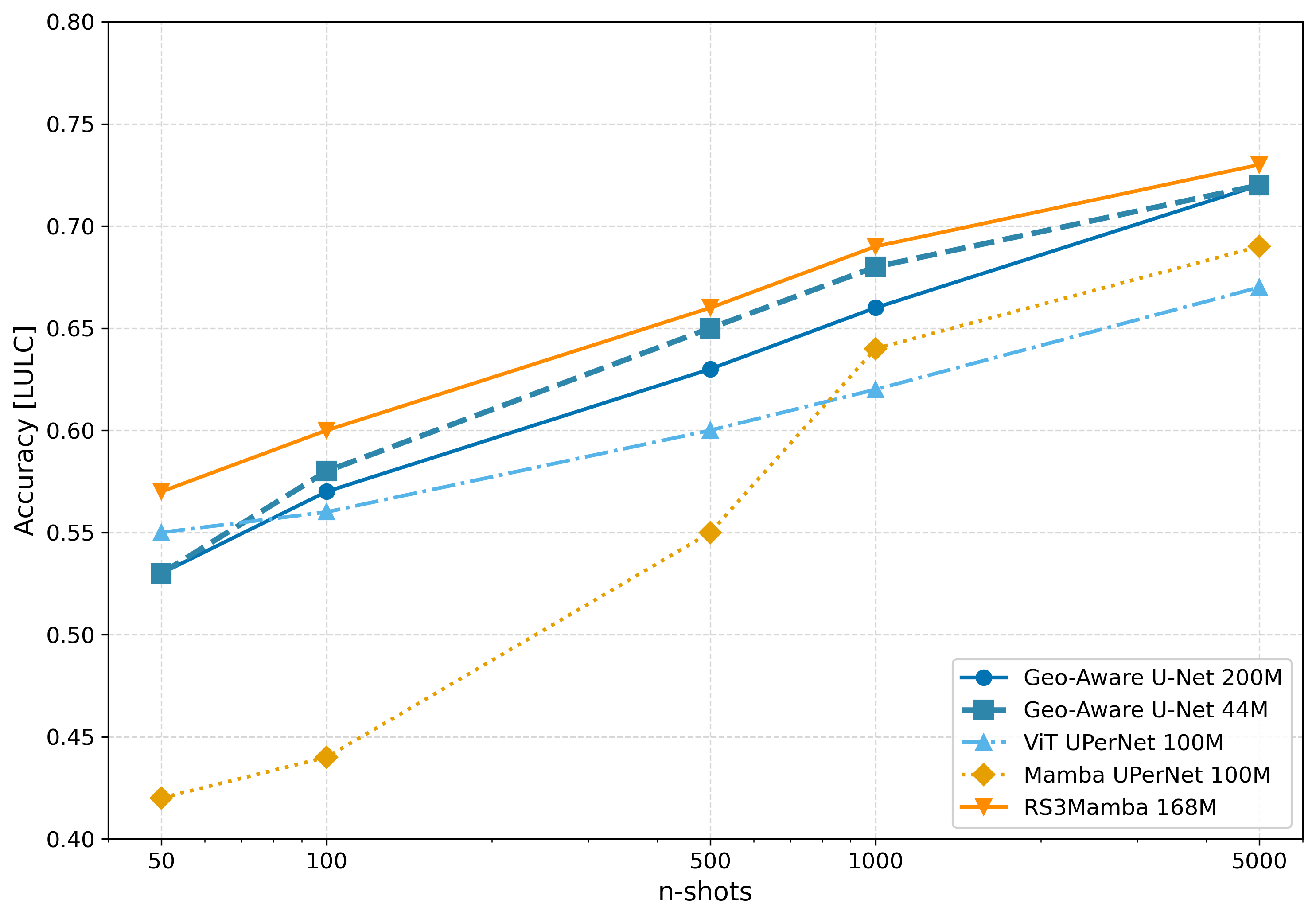}
\caption{Evaluation over PhilEO Bench downstream tasks for various n-shots and for different model pretraining strategies on FastTOM 2TB: Building density regression (top), Road density estimation (middle), and Land cover mapping (bottom), where the key takeaways are that for pixel-wise regression dowsntream tasks, U-Net 200M for most n-shots outperforms the other models, and for semantic segmentation land cover mapping, the RS3Mamba  is effective.} 
\label{fig_2TB}        
\end{figure}

On the 2TB pretraining dataset, the Geo-Aware U-Net 200M model emerges as a strong performer, especially for pixel-wise regression tasks such as roads and buildings. For example, in the building density task with $n=50$, U-Net 200M achieves an RMSE of $0.077$, compared to $0.099$ for the Geo-Aware U-Net 44M. This corresponds to a relative improvement of about $22\%$. More generally, increasing both model size and dataset scale together (Geo-Aware U-Net 200M with 2TB vs. Geo-Aware U-Net 44M with 0.5TB) yields consistent gains. While these improvements may stem from either the larger model or the larger dataset, the results of the Geo-Aware U-Net 44M trained on 2TB provide a way to disentangle these two effects.

For land cover mapping, however, we find that pretraining on the 0.5TB GlobeEO dataset remains advantageous. At $n=100$, the Geo-Aware U-Net 44M pretrained on 0.5TB achieves an accuracy of $0.62$, compared to $0.58$ for the same model pretrained on 2TB, a relative improvement of nearly $7\%$. This effect can be attributed to the seasonal structure of GlobeEO, which samples one year of data uniformly across four seasons. Such temporal diversity aligns closely with the requirements of land cover classification, whereas pretraining on longer temporal sequences, as in MajorTOM 23TB or FastTOM 2TB, relies on random time steps spread over a decade and dilutes this seasonal signal.


Finally, one of the main focal points of this work is the study of Mamba SSM models \cite{paperRS3Mamba,papermamb7}, which offer improved complexity compared to Transformers. Our results suggest that when sufficient labeled data are available, Mamba-based models achieve performance comparable to Transformer-based architectures on pixel-wise regression tasks, narrowing the gap while retaining their computational advantages.
 




%% file: 33_scaling.tex
\subsection{Large-Scale Pretraining (23TB)}\label{sec:Largescale}

\begin{figure}[!t]    
\centering 
\includegraphics[width=3.45in]{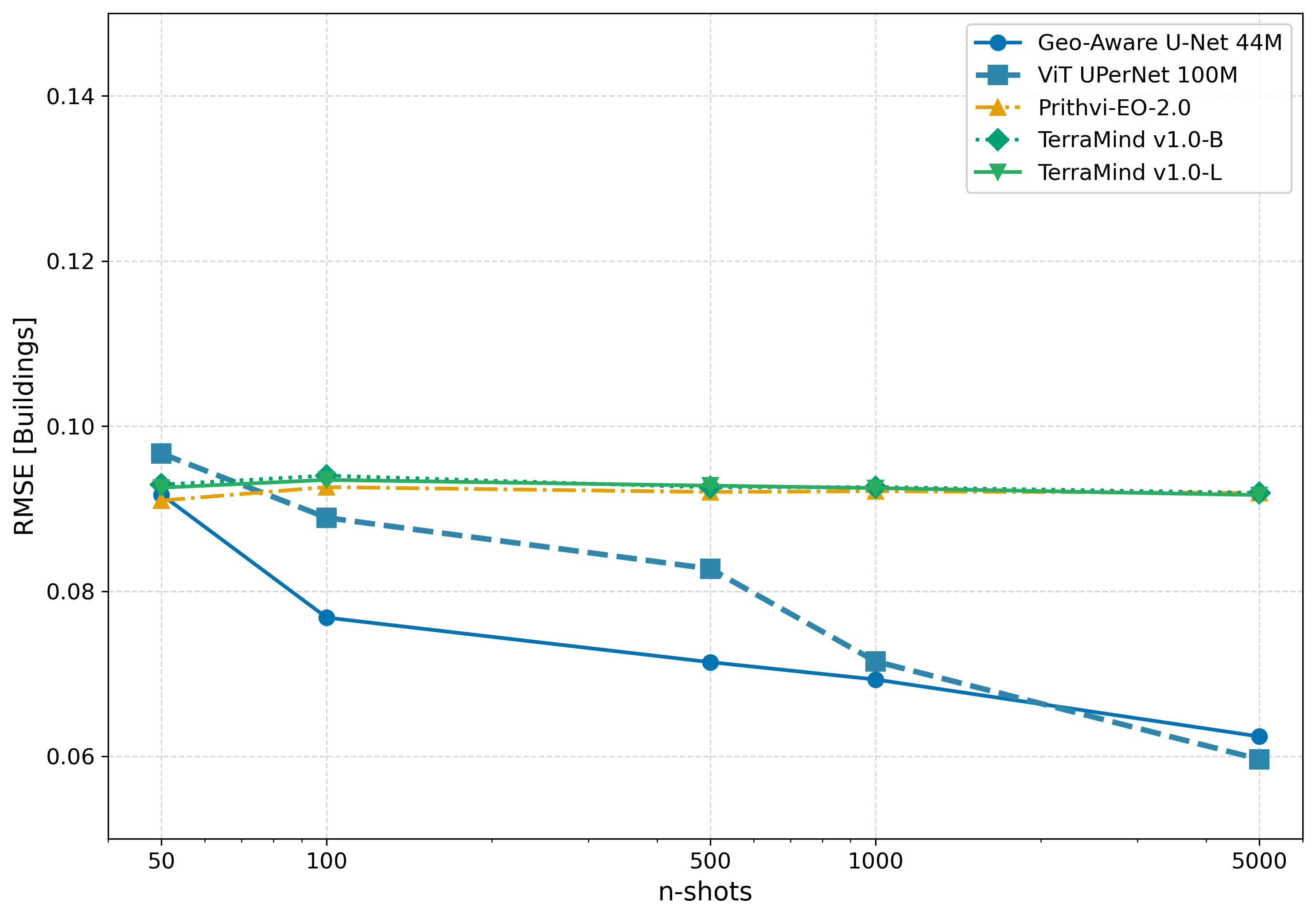}     
\includegraphics[width=3.45in]{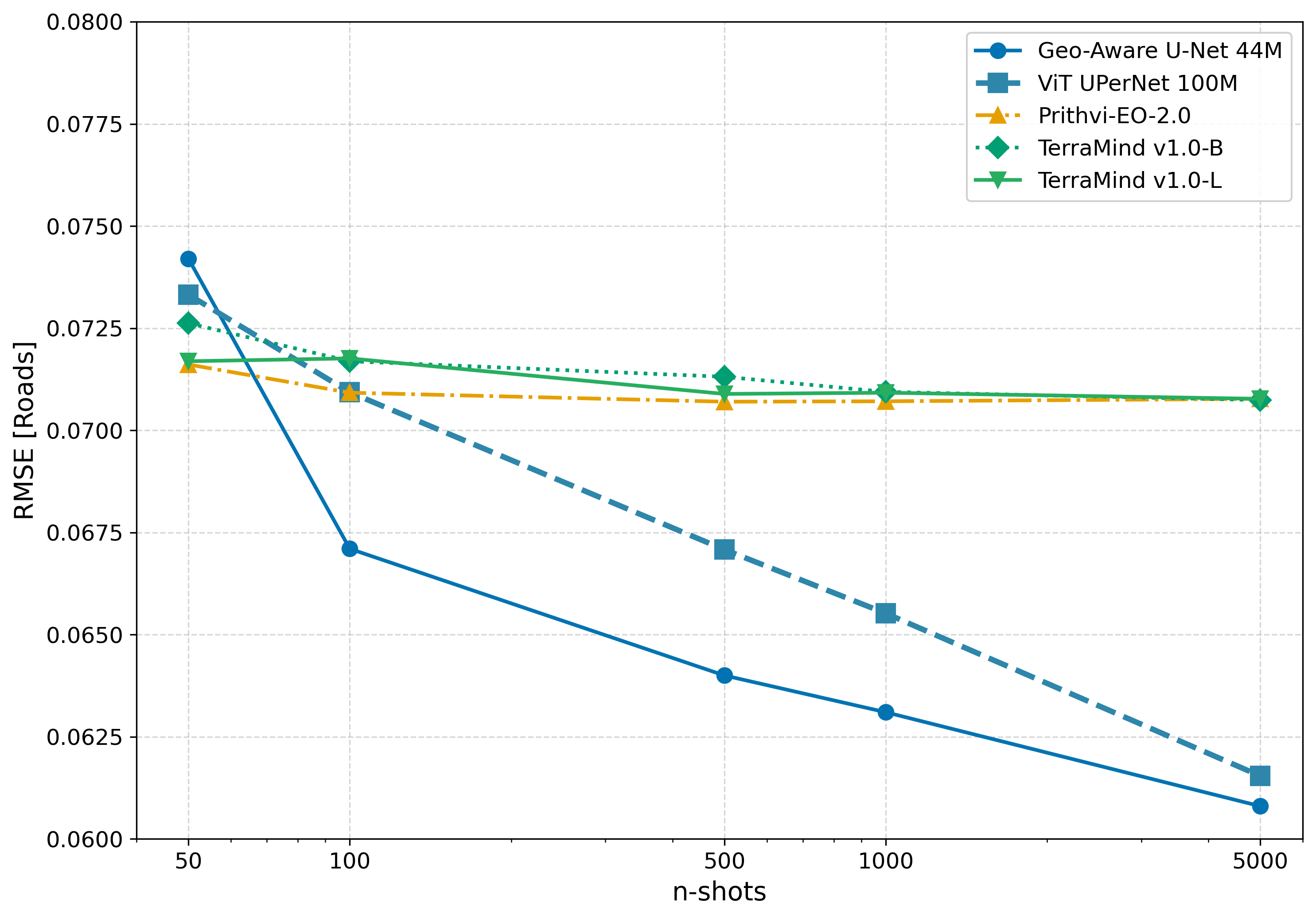}    
\includegraphics[width=3.45in]{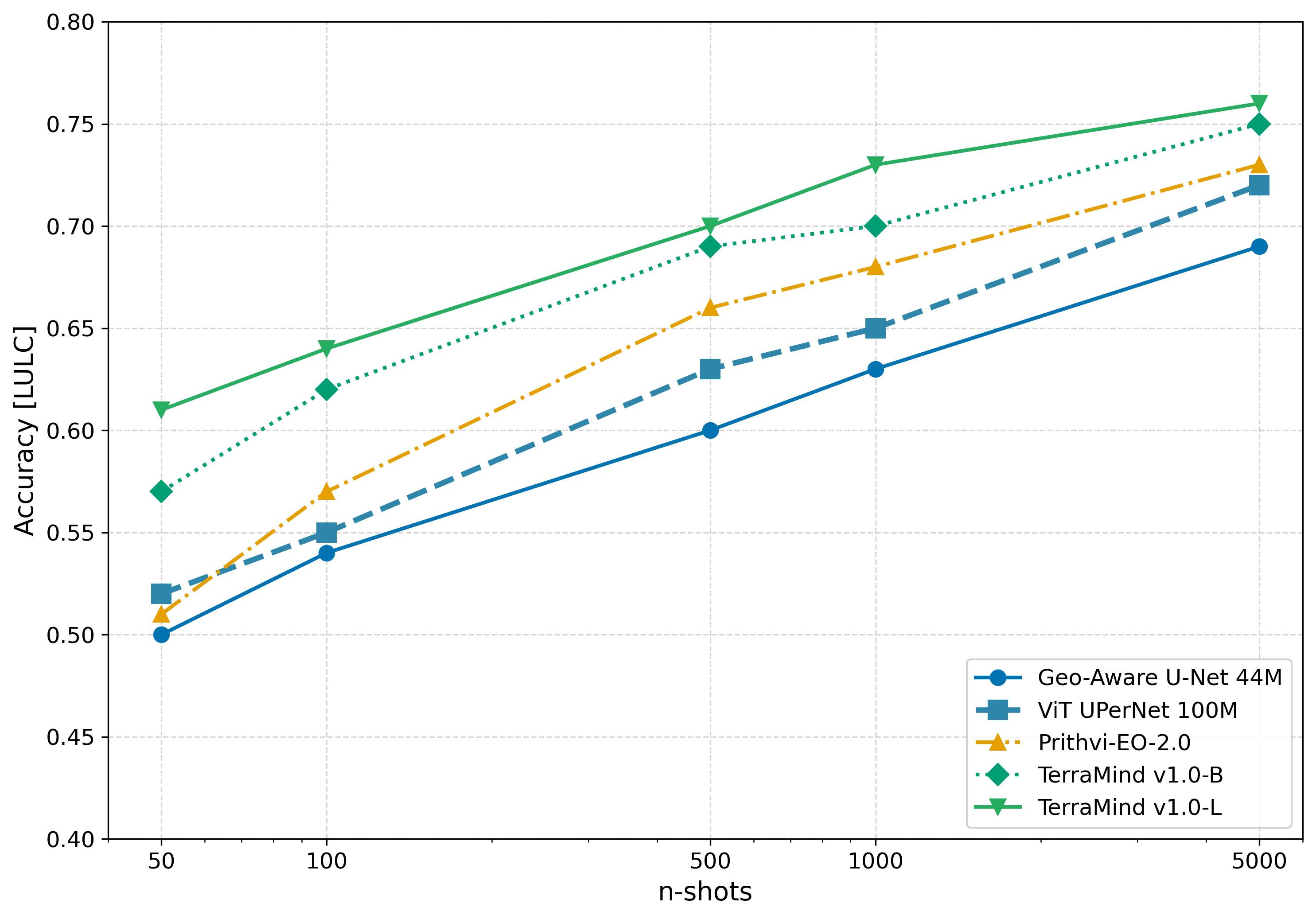}
\caption{PhilEO benchmark results for models pretrained on MajorTOM 23TB: Building density regression (top), Road density estimation (middle), and Land cover mapping (bottom), across $n$-shot settings, where the key takeaways are that for pixel-wise regression tasks, U-Net 44M for most $n$-shots has good performance, and for large $n$-shots, ViT has competitive performance. We also note that for semantic segmentation downstream tasks, ViT UPerNet outperforms U-Net.}   
\label{fig_23TB}       
\end{figure}

We finally scale both model size and pretraining volume using the MajorTOM 23TB dataset (Figure~\ref{fig_23TB}). This large-scale setting is particularly important for evaluating the effect of including diverse geophysical domains such as oceans and ice in pretraining. We compare generalized pretraining on MajorTOM (23TB), which spans land, oceans, ice, and deserts, with specialized pretraining on FastTOM (2~TB), which is restricted to land areas and therefore more closely aligned with the land-focused PhilEO benchmark \cite{phileo}.


For pixel-wise regression tasks such as roads and buildings, U-Net architecture consistently outperforms ViT when both are pretrained on MajorTOM. In the road density estimation task with $n=100$, the Geo-Aware U-Net 44M trained on 23TB achieves an RMSE of $0.067$, compared to $0.071$ for the ViT UPerNet 100M on the same dataset, an improvement of about $5.4\%$. Similar patterns hold for buildings, confirming that CNN-based architectures, with their strong inductive bias for local correlations, remain advantageous in dense regression tasks. The benefits of large-scale pretraining are most visible in the low-shot regimes (below 500 samples), where the Geo-Aware U-Net pretrained on 23TB significantly outperforms smaller-scale baselines, while gains diminish as the number of labeled samples increases.

For semantic segmentation tasks such as land cover mapping, the trend reverses. Here, the ViT UPerNet benefits from its ability to model global correlations and outperforms the U-Net when both are pretrained on MajorTOM. At $n=500$, the ViT UPerNet 100M achieves an accuracy of $0.63$, compared to $0.60$ for the Geo-Aware U-Net 44M, a relative improvement of $5\%$. However, scaling to 23TB does not always translate into better results. At $n=100$, the U-Net 44M pretrained on the smaller GlobeEO 0.5TB dataset achieves an accuracy of $0.62$, whereas both the ViT UPerNet and Geo-Aware U-Net pretrained on 23TB drop to $0.55$ and $0.54$, respectively. In this case, the smaller seasonal dataset outperforms large-scale pretraining by $12.7\%$, underlining the importance of temporally diverse sampling.  

These results suggest that while specialized pretraining incorporates prior knowledge of downstream land-focused tasks, generalized pretraining reflects a broader GFM objective for EO. Including additional domains such as oceans and ice introduces aleatoric uncertainty, but does not necessarily reduce downstream performance on average. Using 23 TB, we increase both the volume and the diversity of the data. For example, in the road task at $n=100$, the Geo-Aware U-Net 44M trained on MajorTOM 23TB achieves $0.067$ RMSE, compared to $0.071$ for the same model trained on GlobeEO 0.5TB, representing a modest $5.6\%$ improvement.

Overall, U-Nets pretrained on MajorTOM continue to outperform ViTs for regression tasks, highlighting the importance of local feature modeling. Yet, for LULC segmentation, GlobeEO remains more effective because of its structured seasonal sampling, which MajorTOM lacks. Related work, such as TerraMind \cite{PaperTerraMind}, has recognized this limitation: the MajorTOM dataset was subsampled and combined with SSL4EO-S12, to create TerraMesh \cite{blumenstiel2025terramesh}, deliberately reducing the share of images from oceans, deserts, and ice to improve downstream relevance.

%% file: 34_FLOPS.tex
\subsection{Computational Efficiency and FLOPs Analysis}\label{sec:flopstuffs}

Beyond downstream task performance accuracy, we also assess computational efficiency, measuring the FLOPs required for each model as a hardware-agnostic estimate of cost \cite{paperforflops}. This is particularly relevant given the quadratic complexity of ViT self-attention, which has motivated many recent works to study model efficiency (e.g., FlashAttention \cite{paperflashatt}, linear or sparse attention \cite{paperlinf,paperlinearattn}, Swin Transformer \cite{swin}, MetaFormer \cite{papermetaf}, and Mamba \cite{papermamb7,papermamb9}).

For a fair comparison, we also report the number of FLOPs required by each model, as recommended in \cite{paperforflops}.  
Lower FLOPs indicate greater computational efficiency.   
FLOPs allow us, together with the final downstream tasks performance results, to decide if it is actually beneficial and worth to scale-up a specific model.      
FLOPs are measured in Python using the DeepSpeed library\footnote{We note that similar values are also obtained using the Python libraries fvcore and calflops.}.    
It is important to note that FLOPs differ from floating-point operations per second (FPS), and we use FLOPs to provide a machine-agnostic measure of computational cost, rather than reporting wall-clock training time.

Looking at the details in Table \ref{tab:maintableevalseeeccc}, we have included models spanning from a very low number of FLOPs, like the Geo-Aware U-Net 44M (with only 5.87G FLOPs) to a very high number of FLOPs, such as ViTs with 300M parameters (more than 500G FLOPs). 

Geo-Aware U-Net 44M model, despite its relatively small size, achieves strong performance with a low computational footprint. This demonstrates that smaller models can outperform larger ones in certain scenarios, particularly for pixel-wise regression tasks.  

When scaling from Geo-Aware U-Net 44M to 200M, FLOPs increase from 5.87G to 82.23G, while RMSE improves modestly. This raises an important question: What is the best model for a given computational cost? And, is there a saturation point beyond which increasing the number of FLOPs yields diminishing returns in performance?

%% file: 35_comparisons.tex
\subsection{Comparative Evaluation with SOTA Architectures}\label{sec:comp}

To better understand the effect of scaling, we benchmarked our approach against two state-of-the-art EO Foundation Models: TerraMind \cite{PaperTerraMind} and Prithvi-EO-2.0 \cite{newibmnasa}. Both represent recent advances in large-scale, general-purpose EO modeling. Prithvi-EO-2.0 is trained on global Harmonized Landsat Sentinel (HLS) archives at unprecedented scale, while TerraMind incorporates multiple modalities—including LULC labels, geo-location, DEM, Sentinel-1, Sentinel-2, and NDVI—into a unified pretraining framework. Their scale and generality make them natural reference points for positioning our work within the emerging landscape of EO Foundation Models. In our experiments, we directly fine-tune their pretrained backbones on the PhilEO benchmark (Figure~\ref{fig_23TB}).

For land cover mapping, TerraMind achieves the strongest performance. At $n=500$, TerraMind v1.0-L reaches an accuracy of $70\%$, compared to $68\%$ for the Geo-Aware U-Net 44M pretrained on GlobeEO 0.5TB, a relative improvement of about $3\%$. This margin is modest but consistent across $n$-shot settings, and can be attributed to TerraMind’s multi-modal pretraining, which encodes priors that directly benefit LULC classification. By contrast, Prithvi-EO-2.0 performs closer to the specialized baselines, confirming that while scaling to global data is advantageous, the use of auxiliary modalities provides the clearest boost in LULC segmentation.

For road and building density estimation, however, the advantage of TerraMind is less pronounced. Improvements remain small even at low $n$-shots, suggesting that its LULC-based pretraining already encodes coarse semantic priors for built environments (e.g., through the “Built” class), but does not transfer as effectively to fine-grained pixel-level regression. In these tasks, CNN-based architectures such as U-Net continue to match or outperform Foundation Model baselines, particularly in the low-data regime, consistent with the patterns discussed in Section~\ref{sec:archicomp}.

Finally, it is worth noting that both TerraMind and Prithvi-EO-2.0 come with significantly higher computational cost compared to the Geo-Aware U-Net 44M, both in terms of parameter count and FLOPs. This highlights the trade-off between the generality of large-scale Foundation Models and the efficiency of specialized architectures.

In summary, while specialized pretraining on smaller, carefully curated datasets such as GlobeEO provides strong baselines, large-scale multi-modal pretraining, as in TerraMind, can achieve competitive or superior performance in semantic segmentation without degrading regression performance. This suggests that EO Foundation Models are beginning to close the gap with specialized approaches, even for land-focused downstream applications.

%% file: 4_discussionconclusion.tex
\section{Conclusion}                           

In this work, we studied the scaling behavior of GFMs across three axes: model architecture, model size, and pretraining data volume. Using the PhilEO benchmark, we compared CNNs, Transformers, and Mamba models across pretraining datasets ranging from $0.5$TB (GlobeEO) to $23$TB (MajorTOM). 
Using $23$TB for pretraining, we do not only increase the size of the data but we also change the data distribution as some classes, such as ice and ocean classes were not represented.
We also contrasted our results with state-of-the-art Foundation Models such as TerraMind and Prithvi-EO-2.0.   

Our experiments reveal three main findings. First, for pixel-wise regression tasks such as road and building density estimation, CNN-based U-Net models remain highly competitive. In particular, the Geo-Aware U-Net 200M achieves the strongest results in low-shot settings, surpassing larger ViT-based models, and highlighting the importance of local correlation modeling in EO imagery. These benefits are most pronounced when scaling both model capacity and pretraining data together (e.g., U-Net 200M with 2TB), while the smaller U-Net 44M also remains surprisingly effective and computationally efficient.

Second, for semantic segmentation tasks such as land cover mapping, ViT UPerNet models provide advantages at scale, benefitting from their ability to capture long-range correlations and leverage multi-level features in the decoder. Nonetheless, specialized pretraining on land-only datasets with seasonal coverage (GlobeEO 0.5TB) yields stronger LULC performance than generalized pretraining on the full 23TB MajorTOM dataset, where ocean and ice imagery introduce temporal and geographical domain shifts. This emphasizes the importance of dataset curation in large-scale EO pretraining.

Third, our evaluation of state-space models (Mamba) shows that they can reach competitive accuracy while being more efficient than Transformers, making them promising candidates for scalable EO modeling. Likewise, our comparison with TerraMind and Prithvi-EO-2.0 demonstrates that while multi-modal large-scale pretraining, as in TerraMind, leads to superior land cover performance ($70\%$ accuracy at $n=500$ vs. $68\%$ for U-Net 44M GlobeEO), the advantages are limited for regression tasks. This suggests that multi-modality enhances semantic segmentation but does not fully replace the need for fine-grained pixel-level learning.       

Overall, our results show that there is no single “best” GFM; indeed, the optimal choice depends on the downstream task and the amount of available labeled data. U-Net models excel in low-shot regression, ViTs benefit semantic segmentation at scale, and Mambas offer a balanced trade-off between accuracy and efficiency. Importantly, careful dataset design, particularly in terms of temporal sampling and domain relevance, can be as critical as model architecture or size.

As GFMs continue to scale, our findings underscore the need for systematic evaluation across multiple axes of variation, including architecture, dataset composition, and computational efficiency, reconsidering a paradigm shift from model-centric to data-centric DL \cite{datacentricml}. Future work should extend this study to multi-temporal and multi-modal pretraining, as well as explore knowledge distillation and on-board deployment scenarios where FLOPs efficiency is essential.

%% file: 5-annexMamba.tex
\section*{Appendix}  
\section*{Further experiments using Mamba SSM}

\begin{figure}[!t]                                   
\centering              
\includegraphics[width=3.55in]{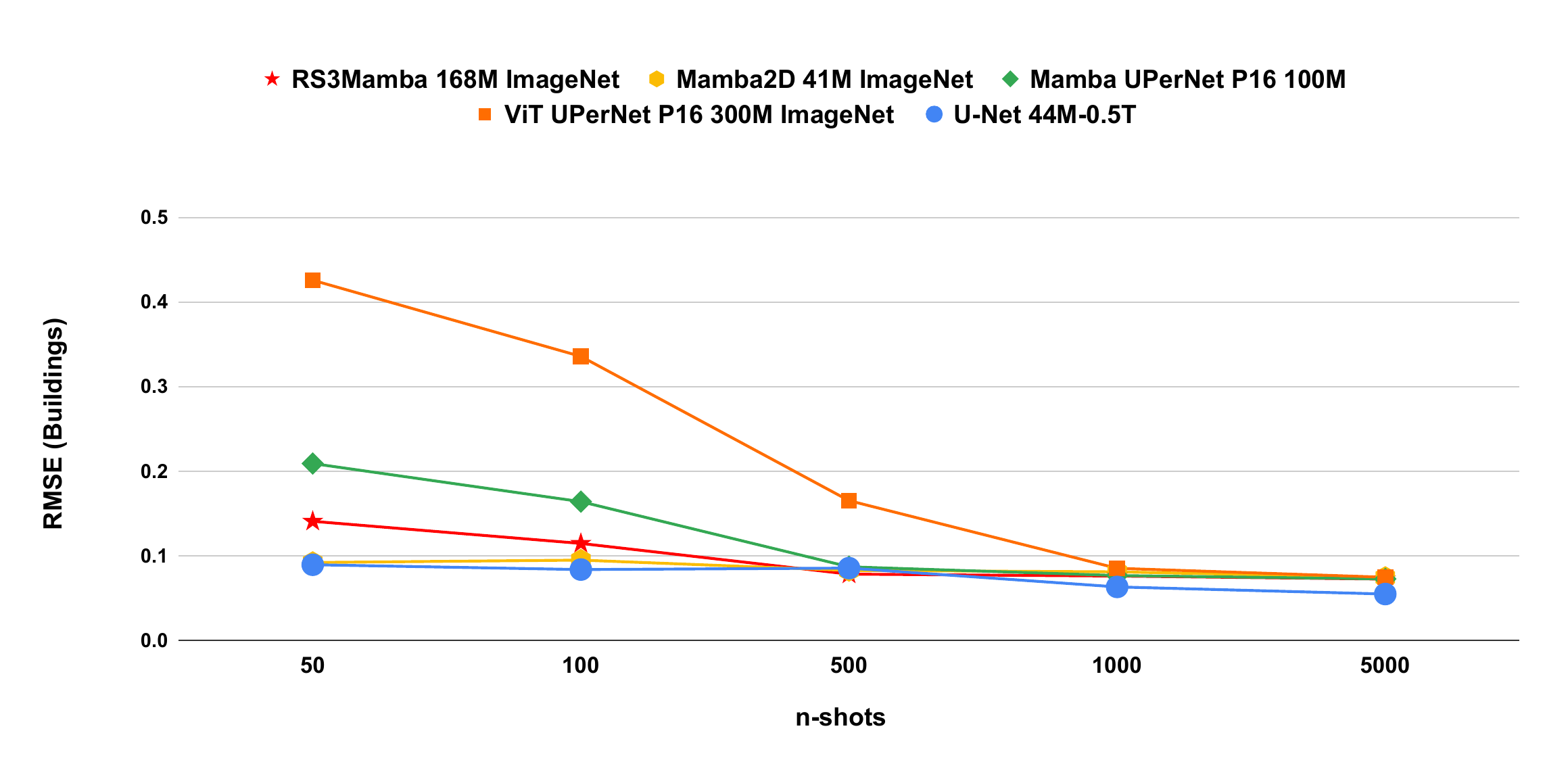}     
\includegraphics[width=3.55in]{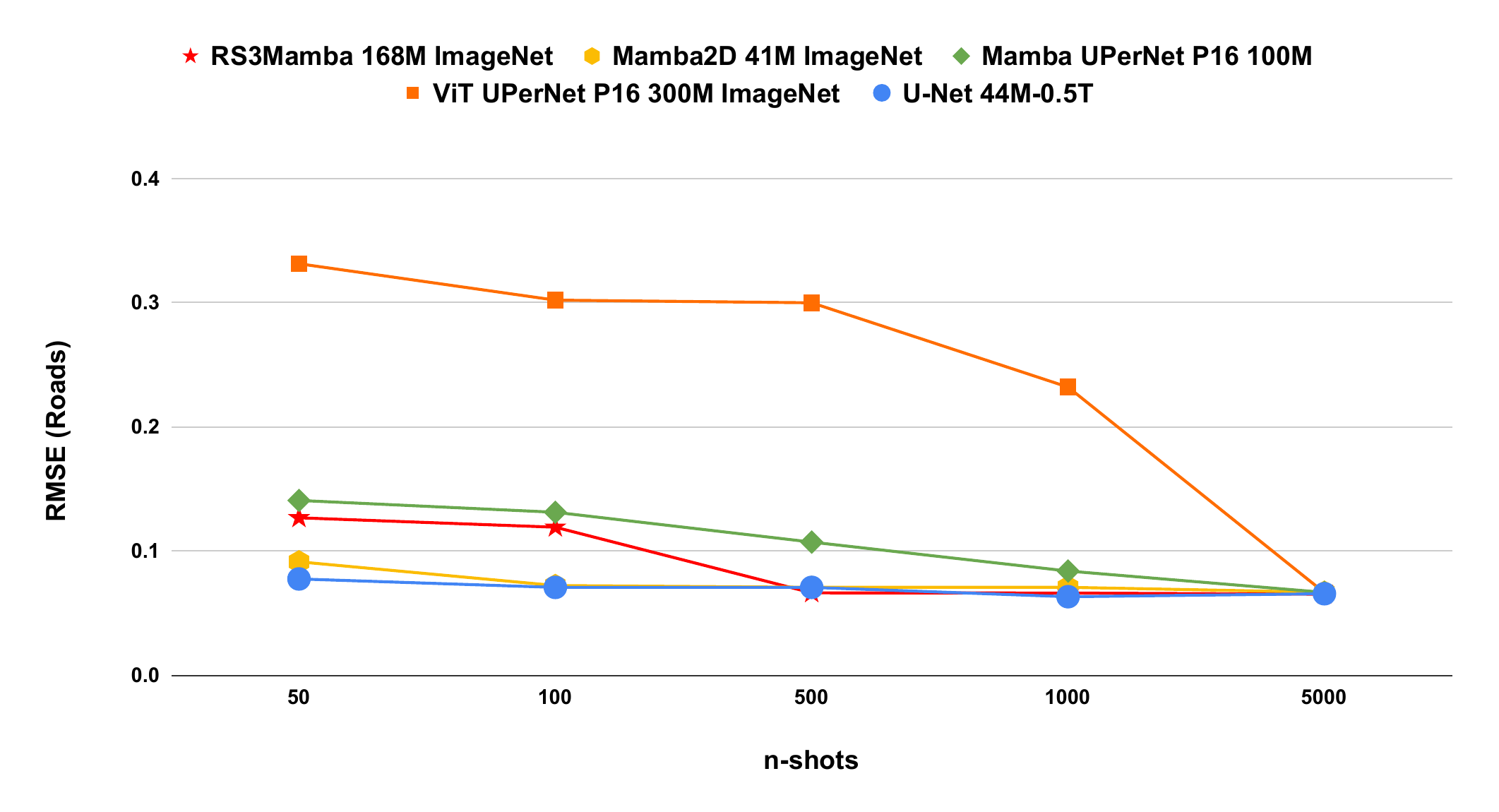}     
\includegraphics[width=3.55in]{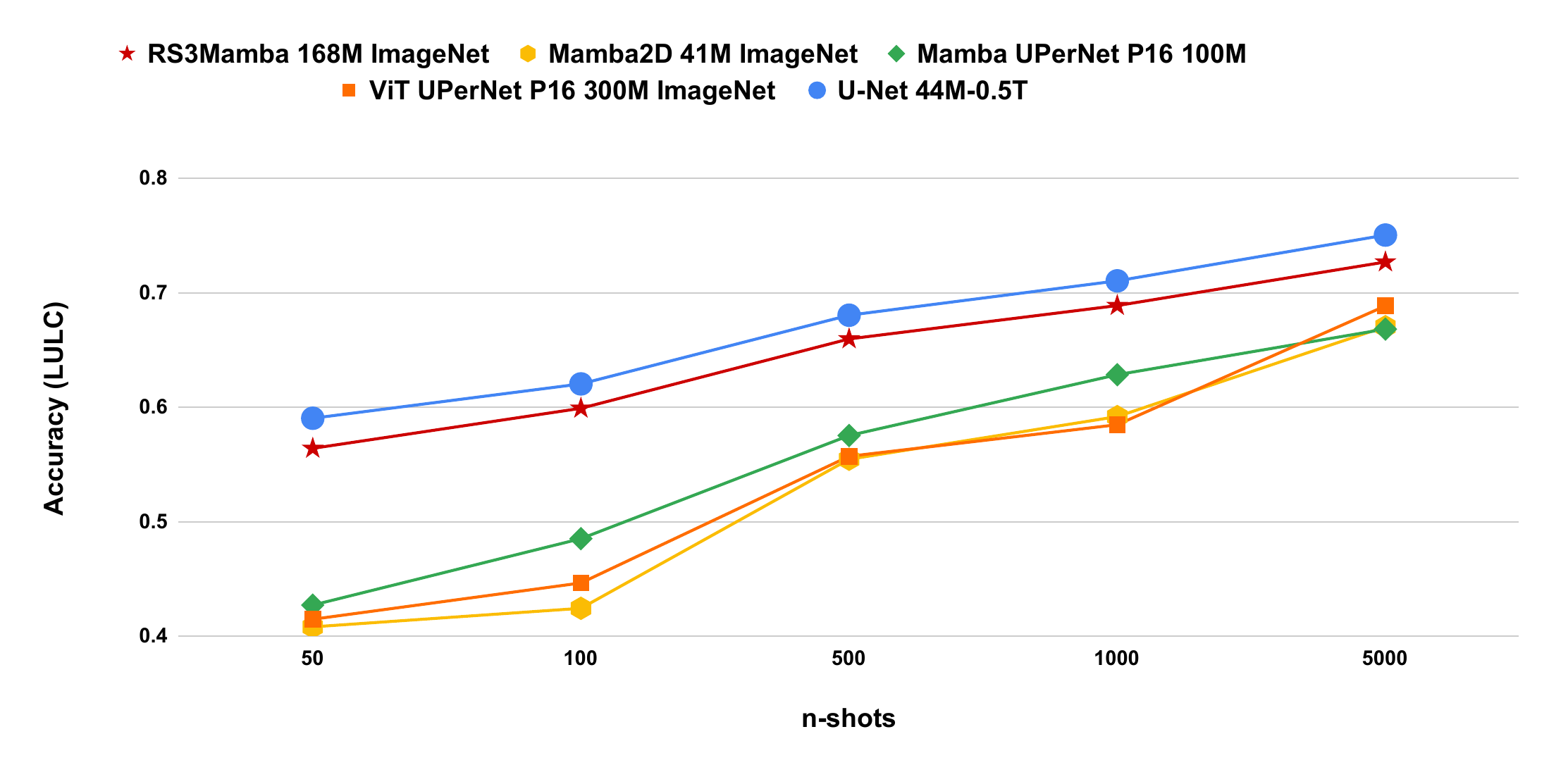}
\caption{Further evaluation and experiments of Mamba SSM models over PhilEO Bench downstream tasks for various $n$-shots for various models pretrained on ImageNet: Building density regression (Top), Road density estimation (Middle), and Land cover mapping (Bottom).}   
\label{fig_furthMamb}      
\end{figure}

We did further experiments using Mamba SSM models and for this set of results, the models are pretrained on the ImageNet dataset.
The size of ImageNet is $130$GB and it is \textit{labelled}, i.e. supervised learning is performed (rather than self-supervised learning).
Figure \ref{fig_furthMamb} shows the results for this set of experiments. 

The main findings are that for limited labelled data, it is beneficial to pretrain on Earth Observation satellite data rather than on ImageNet, and that Mamba models like RS3Mamba \cite{paperRS3Mamba} can be effective for semantic segmentation land cover mapping.  

